\pdfoutput=1
\documentclass{llncs}

\usepackage{makeidx, graphicx}  % allows for indexgeneration
\usepackage{amsfonts}
\usepackage{booktabs}
\usepackage{siunitx}
\usepackage{filecontents}
\usepackage{soul}

\usepackage[usenames]{color}
\definecolor{grey}{rgb}{0.33, 0.33, 0.33}
\definecolor{red}{rgb}{1,0,0}
\definecolor{blue}{rgb}{0,0,1}
\newcommand\mdoubleplus{\mathbin{+\mkern-10mu+}}
\begin{document}
\frontmatter          % for the preliminaries
\pagestyle{headings}  % switches on printing of running heads
\addtocmark{Hamiltonian Mechanics} % additional mark in the TOC
\mainmatter              % start of the contributions
\title{Can Deep Learning Relax Endomicroscopy Hardware Miniaturization Requirements?}
\titlerunning{Deep Learning Super-Resolution for Endomicroscopy}  % abbreviated title (for running head)

%
\authorrunning{Izadi et al.} % abbreviated author list (for running head)
% \tocauthor{anonymous}
%
\author{Saeed Izadi,  Kathleen P. Moriarty, and Ghassan Hamarneh}
\institute{
School of Computing Science, Simon Fraser University, Canada \\
\email{\{saeedi, kmoriart, hamarneh\}@sfu.ca}\\
}

\maketitle              % typeset the title of the contribution

\begin{abstract}
Confocal laser endomicroscopy (CLE) is a novel imaging modality that provides \textit{in vivo} histological cross-sections of examined tissue. Recently, attempts have been made to develop miniaturized \textit{in vivo} imaging devices, specifically confocal laser microscopes, for both clinical and research applications. However, current implementations of miniature CLE components, such as confocal lenses, compromise image resolution, signal-to-noise ratio, or both, which negatively impacts the utility of \textit{in vivo} imaging. In this work, we demonstrate that software-based techniques can be used to recover lost information due to endomicroscopy hardware miniaturization and reconstruct images of higher resolution. Particularly, a densely connected convolutional neural network is used to reconstruct a high-resolution CLE image from a low-resolution input. In the proposed network, each layer is directly connected to all subsequent layers, which results in an effective combination of low-level and high-level features and efficient information flow throughout the network. To train and evaluate our network, we use a dataset of 181 high-resolution CLE images. Both quantitative and qualitative results indicate superiority of the proposed network compared to traditional interpolation techniques and competing learning-based methods. This work demonstrates that software-based super-resolution is a viable approach to compensate for loss of resolution due to endoscopic hardware miniaturization.

%\keywords{confocal laser endomicroscopy, hardware miniaturization, super-resolution, image reconstruction, dense convolutional neural networks, deep learning}
\end{abstract}
\section{Introduction}
%
%\km{GI cancer is .... + statistics \newline}

Last year, colorectal cancer caused an estimated 50,260 deaths in the United States alone and another 140,030 people are expected to be diagnosed with this disease during 2018~\cite{siegel2017colorectal,siegel2018cancer}. Accordingly, it is the third most commonly diagnosed cancer among both men and women~\cite{siegel2018cancer}. Early diagnosis and treatment of colorectal cancer is crucial for reducing the mortality rate.  Gastroenterologists screen and monitor the status of their patients' digestive systems through specialized endoscopy procedures such as colonoscopy and sigmoidoscopy. During colonoscopy, a flexible video endoscope is guided through the large intestine, capturing images used to differentiate between neoplastic (intraepithelial neoplasia, cancer) and non-neoplastic (e.g., hyperplastic polyps) tissues. \par

% \km{lead to CLE modality \newline}
Since the introduction of endoscopy to gastroenterology, many significant advances have been made toward improving the diagnostic and therapeutic yield of endoscopy. Confocal laser endomicroscopy (CLE), first introduced to the endoscopy field in 2004~\cite{kiesslich2004confocal}, is an emerging imaging modality that allows histological analysis at cellular and subcellular resolutions during ongoing endoscopy. An endomicroscope is integrated into the distal tip of a conventional video colonoscope, providing an \textit{in vivo} microscopic visualization of tissue architecture and cellular morphology in real-time. Endomicroscopes offer a magnification and resolution comparable to that obtained from \textit{ex vivo} histology imaging techniques, without the need for biopsy (i.e., tissue removal, sectioning and staining). \par

Despite the promise of confocal laser endomicroscopy, both clinicians and researchers prefer compact instruments with relatively large penetration depth to recognize tissue structures such as the mucosa, the submucosa, and the muscular layers. Compact instruments can also directly benefit the patients, as smaller devices improve early diagnostic procedures by offering greater flexibility during hand-held use, for a quicker and less invasive endoscopy~\cite{helmchen_2002}. In this regard, further attempts have been made to design miniaturized confocal scanning lasers capable of capturing images from the tissue subsurface with micron resolution \textit{in vivo}, once installed on top of a flexible fiber bundle. However, miniaturization implies using smaller optical elements, which introduces pixelation artifacts in images. Therefore, there exists a trade-off between miniaturizing the CLE components and the resultant image resolution. \par

Image super-resolution, transforms an image from low-resolution (LR) to high-resolution (HR) by recovering the high-frequency cues and reconstructing textural information. In the past decade, various learning-based approaches have been proposed to learn the desired LR-to-HR mapping, including dictionary learning~\cite{5466111,Yang2012}, linear regression~\cite{6751179,timofte2014a+}, and random decision forests~\cite{7299003}.\par

In recent years, deep learning models have been applied to various image interpretation tasks. Among such efforts, convolutional neural networks (CNN) have been utilized to resolve the ill-posed inverse problem of super-resolution. Dong et al.~\cite{Dong2016srcnn} demonstrated that a fully convolutional network trained end-to-end can be used to perform the LR-to-HR nonlinear mapping. The same authors extended their previous work by introducing deconvolutional layers at the end of the architecture, such that the mapping between LR and HR images is learned directly without image interpolation~\cite{dong2016accelerating}. They also slightly increased the depth of the network and adopted smaller kernels for better performance. Instead of HR images, Kim et al.~\cite{Kim_2016_VDSR} suggested to train deeper neural networks through predicting the residual images, which when summed with an interpolated image gives the desired output. Increasing the network depth by adding weighted layers introduces more parameters, which can lead to overfitting. Kim et al.~\cite{Kim2016_rCNN} tackled overfitting by using a deeply-recursive convolutional network. In their work, the same convolutional layers are used recursively without the need for extra parameters. To simplify the training of the network, they suggested recursive supervision and skip connections to avoid the problem of vanishing/exploding gradients. \par

Given the constraints imposed by CLE hardware miniaturization, we propose to leverage state-of-the-art deep learning super-resolution methods to mitigate the unwanted trade-off between miniaturization and image resolution. In other words, we show that the pixelation artifact, which is a consequence of hardware miniaturization, can be significantly remedied through an efficient and practical use of software-based techniques, particularly machine learning methods. To this end, we employ a densely connected CNN in which extensive usage of skip connections is exploited~\cite{Tong2017_DLSR}. Dense connections help information flow in backpropagation algorithms and alleviate the vanishing gradient problem. Furthermore, the low-level features from early layers are efficiently combined with those of later layers. In addition, we use sub-pixel convolutional layers~\cite{Shi2016_subpixCNN} to render the upsampling operation learnable and expedite the reconstruction process.   

\section{Method} 
\begin{center}
\begin{figure*}[t!]
  \includegraphics[width=\textwidth]{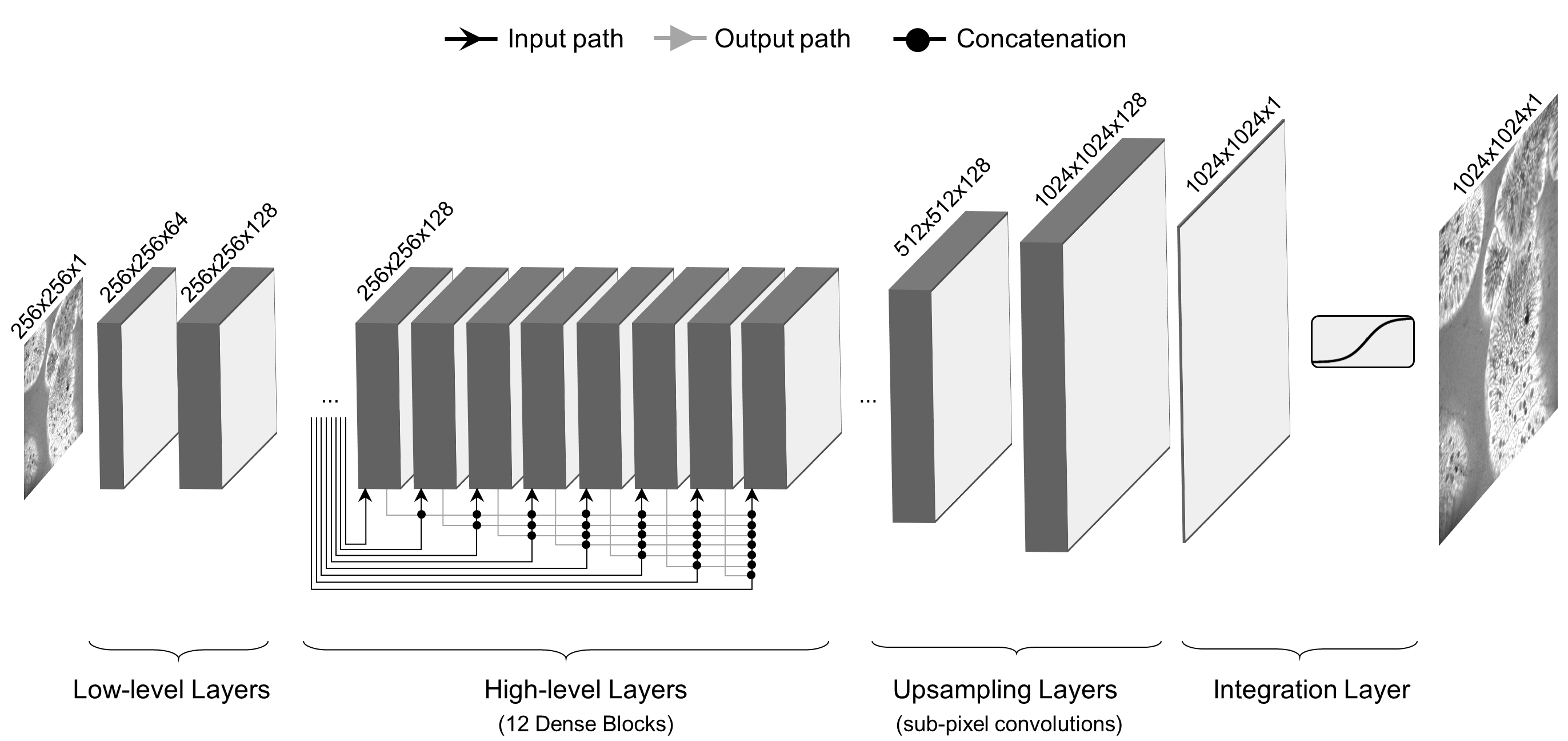}
  \caption{Overall architecture of the DenseNet model, shown here for $\times4$ scale factor, i.e., from a 256$^2$ LR input image to a 1024$^2$ HR output image. In each dense block, convolutional layers are connected to all subsequent layers.}
  \label{fig:overall_arch}
\end{figure*}
 \end{center}

Our main goal in this work is to super-resolve an LR image by passing it through a set of nonlinear transformations to recover high-frequency details and reconstruct the HR image, effectively increasing the number of pixels from $N_{LR}\times N_{LR} \text{ to } N_{HR}\times N_{HR}$, where $\frac{N_{HR}}{N_{LR}}$ is the scale factor. The proposed architecture consists of dense blocks and upsampling layers which are efficiently designed to combine the features from earlier layers with those of later layers and improve information flow throughout the model. Fig.~\ref{fig:overall_arch} depicts the architecture of the employed model.\par
\noindent \textbf{Low-Level Features}. A series of low-level features are extracted from small regions of the LR input image using two successive convolutional layers with kernel size $3\times3$ and ReLU non-linearity. The number of feature channels for the first and second layer is 64 and 128, respectively. The learned low-level features are used to efficiently represent the intrinsic textural differences between LR and HR images. \par
\vspace{-12mm}
\noindent \textbf{High-Level Features}. The resultant low-level feature maps are used as the input to a fully convolutional DenseNet architecture to provide high-level features. DenseNet, which was first introduced by Huang et al.~\cite{Huang2017_denseCNN}, consists of a set of dense blocks in which any layer is connected to every other layer in a feed-forward fashion. Alternatively stated, the $i^{th}$ layer in a dense block receives the concatenation of outputs by all preceding layers as the input:
\begin{equation}
   L_i = relu(\psi_{\theta^i}(L_1 \mdoubleplus  L_2 \mdoubleplus  ... \mdoubleplus  L_{i-1}))
\end{equation}
where $\psi_{\theta^i}$ denotes the transformation of  the $i^{th}$ layer parameterized by $\theta^i$ and $\mdoubleplus$ denotes the concatenation operation. Dense skip connections help alleviate the vanishing-gradient problem and improve information flow throughout the network. Counter-intuitively, the number of parameters is also reduced since the previously-generated feature maps are re-used in the subsequent layers, thus minimizing the need for learning redundant features. As depicted in Fig.~\ref{fig:overall_arch}, a single dense block consists of $m$ convolutional layers, each producing $k$ feature maps, referred to as the \textit{growth rate}. Accordingly, the final output of each dense block has $m \times k$ features maps. The growth rate regulates how much new information each layer contributes to achieving the final performance. In this study, we set $m$ and $k$ to be 8 and 16, respectively. Thus, each dense block receives and produces 128 feature maps as input and output. We stack 12 dense blocks in a feed-forward fashion to construct the DenseNet part of our proposed architecture.

\noindent \textbf{Upsampling Layers}. In some SR methods~\cite{7410407,Dong2016srcnn,Kim_2016_VDSR}, the LR image is first resized to match the HR spatial dimensions using bicubic interpolation. Thereafter, several convolution layers are employed to enhance the interpolated input in the HR space. In addition to having a considerable increase in memory usage and computational complexity, these interpolation methods are categorized as non-learnable upsampling techniques, which do not leverage data statistics to bring new information for more accurate reconstruction. As an alternative, deconvolutional layers, which are learnable operations, are utilized to enlarge the spatial dimensions of the LR image. However, the most prominent problem associated with deconvolutional layers is the presence of checkerboard artifacts in the output image. To overcome this, extra post-processing steps or smoothness constraints are required. In this work, we use sub-pixel convolutional layers~\cite{Shi2016_subpixCNN}, to upsample the spatial size of the feature maps within the network. 
Suppose that we desire to spatially upsample $c$ feature maps of size $h \times w \times c$ to size $H \times W \times c$, 
by a scale factor $r=H/h=W/w$. 
The LR feature maps would be fed into a convolution layer that increases the number of channels by a factor of $r^2$, resulting in a volume of size $h \times w \times (c \times r^2)$. Next, the resultant volume is simply re-arranged to be of shape $(h\times r) \times (w \times r) \times c$, which is equal to $H \times W \times c$. Here, we use successive $\times 2$ upsampling layers to gradually increase the spatial dimensionality. Each upsampling block contains a single convolutional layer with $3 \times 3$ kernel size and ReLU non-linearity.\par

\noindent \textbf{Integration Layer}. Once the features maps match the spatial dimension in the HR space, an integration layer is used to consolidate the features across the channels into a single channel. The integration layer is a convolutional layer with $3 \times 3$ kernel size and a single output channel. Finally, a \textit{sigmoid} activation function is employed to produce the super-resolved image.

\section{Experiments}
\noindent \textbf{Data}. We evaluate our study on the dataset provided by Leong et al.~\cite{LEONG20081870}. The dataset contains 181 gray scale confocal images of size $1024 \times 1024$ from 31 patients and 50 different anatomical sites. Each patient has undergone a confocal gastroscopy (Pentax EC-3870FK, Pentax, Tokyo, Japan) under conscious sedation. CLE images and forceps biopsies of the same sites were taken sequentially at standardized locations (i.e., sites of the small intestine). Each forceps biopsy was then assessed by 2 experienced blinded histopathologists. Despite our application of interest being colorectal cancer, we used the publicly available CLE celiac dataset as a proof-of-concept. Colorectal cancer images are assessed primarily in the large intestine as opposed to the small intestine used in celiac assessment, however the imaging procedure (CLE) remains the same. This dataset was made publicly available as part of an International Symposium on Biomedical Imaging (ISBI) challenge %~\cite{isbi2017_challengedata},
and we used the provided training and test sets, consisting of 108 and 73 images, respectively.\par

\noindent \textbf{Implementation Details}. We partition the HR images into $64\times64$ non-overlapping patches. Then, the HR patches are downsampled by bicubic interpolation to construct $<$LR, HR$>$ pairs for training the model. The network is optimized with Adam~\cite{kingma2014adam} optimizer with default parameters, i.e. $\beta_1=0.9$, $\beta_2=0.999$ and $\epsilon=10^{-4}$. We set the mini-batch size to 128. The learning rate is first initialized with 0.001 and is multiplied by $\gamma=10$ at epochs 50 and 200. The network is trained for 300 epochs using L1 loss. For data augmentation, we  use random horizontal and vertical flips. The proposed method is implemented in PyTorch and is trained using two Nvidia Titan X (Pascal) GPUs. It takes 2 days to train the networks for each upsampling factor. All hyper-parameters (optimizer, learning rate, batch size, and distance metric) are found via grid search on 20 images from the training set. \par
\begin{figure*}[!t]
  \includegraphics[width=\textwidth]{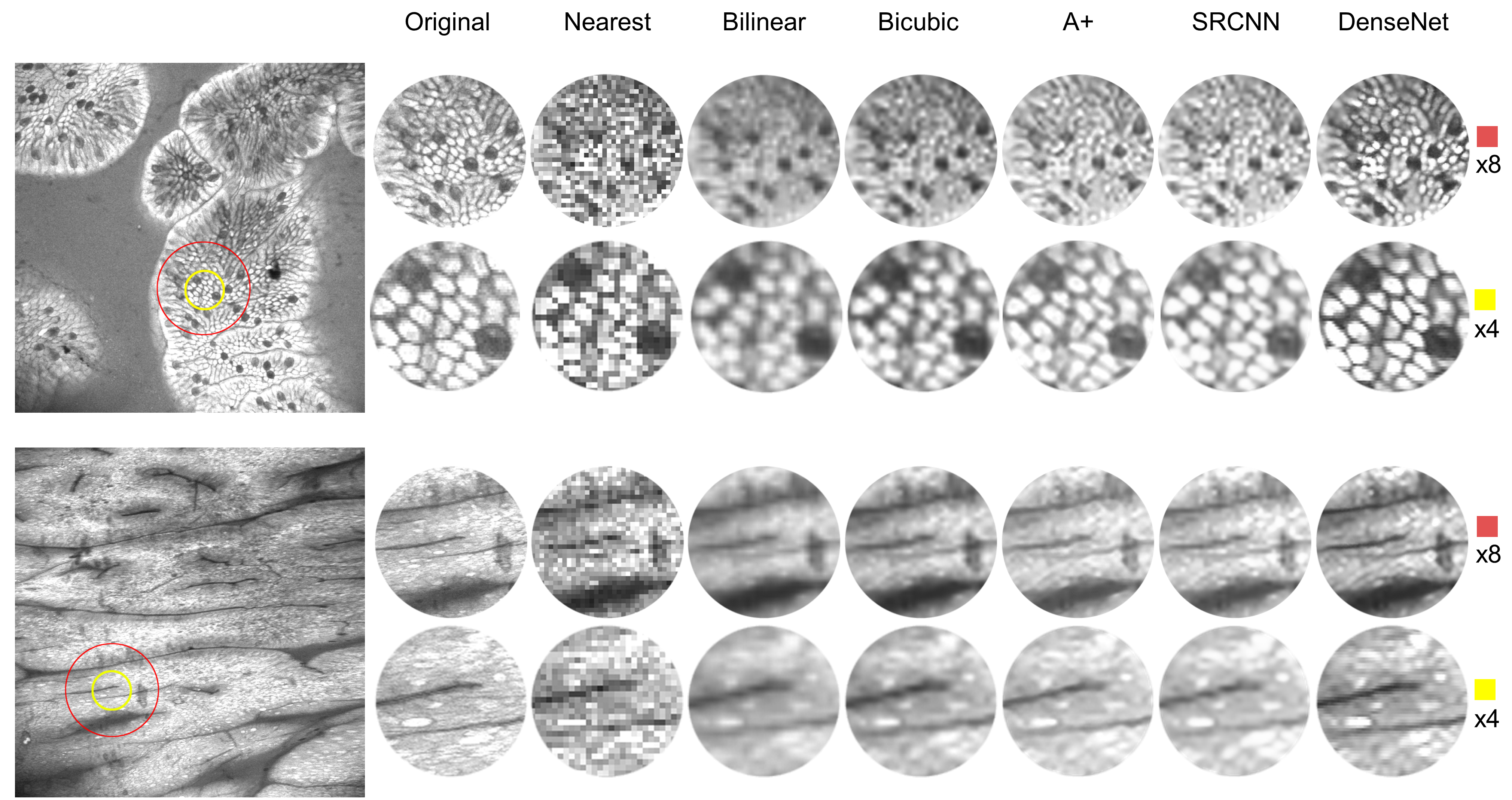}
  \caption{Qualitative Results for two sample images. For each image, the first and second circle rows show the zoomed-in patches for $\times4$ and $\times8$, respectively. }
  \label{fig:qualitative}
\end{figure*}
\noindent \textbf{Qualitative Results}. In Fig.~\ref{fig:qualitative}, we visually compare our proposed super-resolution method to three traditional interpolation techniques and two learning-based approaches with scale factors of $\times 2$, $\times 4$ and $\times 8$. Evidently, DenseNet produces output images of higher quality by reconstructing high-frequency cues and removing visual artifacts, e.g. over-smoothness and pixelation. Specifically for a $\times 8$ scale factor, the densely connected network can accurately recover high-level textural patterns such as grids and granular patterns. Moreover, a more rigorous examination of smaller regions for $\times4$ scale factor clearly reveals the superiority of DenseNet model in producing sharper edges and improved contrast for lines and shapes. \par

\begin{center}
\begin{figure*}[t!]
  \includegraphics[width=\textwidth]{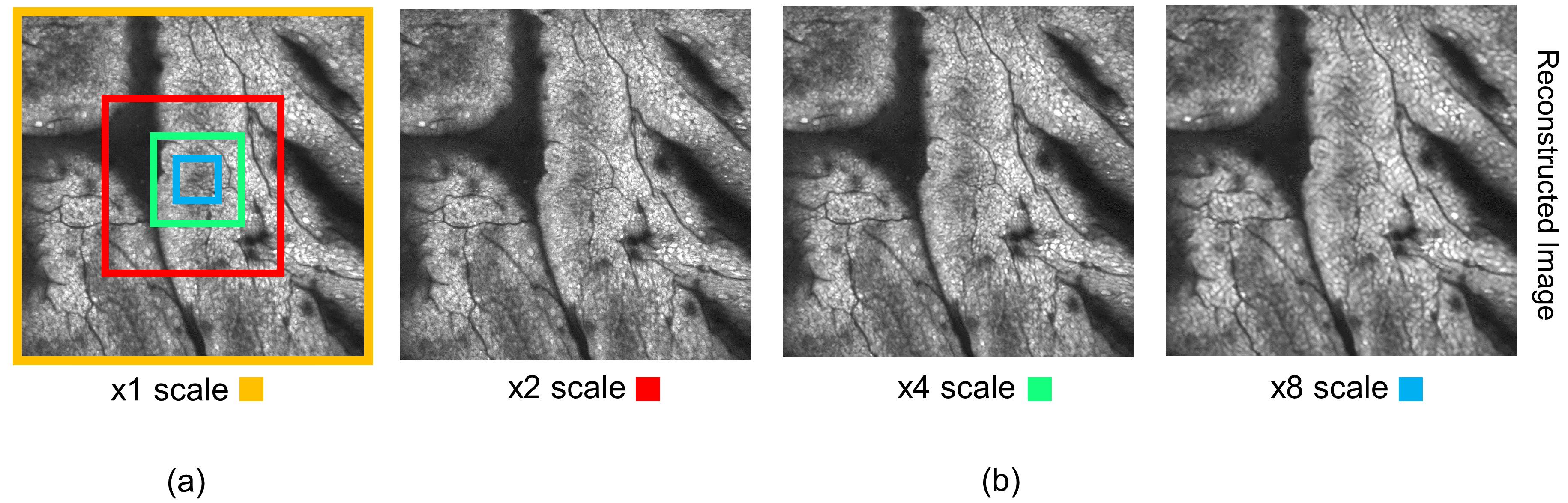}
  \caption{Reconstruction analysis. (a) visualizes the amount of lost pixels for different scale factors relative to the original size. (b) shows the reconstructed images for scale factors $\times2, \times 4$ and $\times 8$.}
  \label{fig:reconstruct}
\end{figure*}
\end{center}

From a clinician's point of view, the reconstruction power of the method offers a clear advantage over others. In Fig.~\ref{fig:reconstruct} we illustrate the trade-off between the amount of lost information after downsampling and the quality of the reconstructed image. As can be seen, a large portion of pixels is discarded in downsampling, restricting the networks to a small fraction of the original image pixels for reconstruction. However, deep learning approaches are clearly capable of generating a sharp image from only 1.6\% of pixels (for a scale factor of $\times 8$)  with very small L1 distance values which indicates a minimal loss of information. \par

\noindent \textbf{Quantitative Results}. Table.~\ref{tab:quant} compares our proposed method with three interpolation methods and two learning-based techniques in terms of PSNR (Peak Signal to Noise Ratio) and SSIM (Structural Similarity). PSNR is a well-known metric for image quality assessment which is inversely proportional to Mean Square Error. SSIM also measures the the similarity between two images and is correlated with quality perception in human visual system. In terms of PSNR, DenseNet yields $2.08$, $1.93$ and $1.14$ average improvements over Nearest, Bilinear and Bicubic interpolation methods across all scale factors, respectively. For learning-based approaches, DenseNet outperforms A+~\cite{timofte2014a+} and SRCNN~\cite{Dong2016srcnn} in terms of average SSIM by $0.020$ and $0.019$ over all scale factors, respectively.

\begin{table}[t!]
    \centering
    \begin{tabular*}{\textwidth}{c
    @{\extracolsep{\fill}}ccccccccccccc} \toprule
    
    \multicolumn{1}{c}{} & \multicolumn{2}{c}{\textbf{Nearest}} & \multicolumn{2}{c}{\textbf{Bilinear}} & \multicolumn{2}{c}{\textbf{Bicubic}} &\multicolumn{2}{c}{\textbf{A+}} &\multicolumn{2}{c}{\textbf{SRCNN}} &\multicolumn{2}{c}{\textbf{DenseNet}} \\
& {PSNR}  & {SSIM} & {PSNR} & {SSIM}  & {PSNR} & {SSIM} & {PSNR} & {SSIM} & {PSNR} & {SSIM}  & {PSNR} & {SSIM} \\ \midrule
    \textbf{$\times$2}  & 35.32  & 0.881 & 34.21  & 0.849  & 35.80  & 0.908  & 36.21 & 0.925 & 35.54 & 0.930 & \textbf{38.57}  & \textbf{0.950} \\
    \textbf{$\times$4}  & 31.64  & 0.658 & 32.38  & 0.707  & 32.87  & 0.755 & 33.00 & 0.781 & 33.01 & 0.778 & \textbf{33.32}  & \textbf{0.801} \\
    \textbf{$\times$8}  & 30.59  & 0.528 & 31.40  & 0.586  & 31.70  & 0.615& 31.74 & 0.636 & 31.80 & 0.636 & \textbf{31.90}  & \textbf{0.651} \\ \bottomrule
    \vspace{0.5pt}
    \end{tabular*}
    \caption{Quantitative Results. Average PSNR and SSIM scores for scale factors $\times2, \times 4$ and $\times 8$ on 73 test images. } 
    \label{tab:quant} 
\end{table}
\vspace{-14mm}
\section{Conclusion}
Developing smaller hardware for medical imaging  devices has several advantages such as increased portability and reduced patient discomfort. However, hardware miniaturization comes at the expense of reduced image quality. In this preliminary study, we obtained encouraging results to support that software-based methods can be used to counteract the loss of image quality due to miniaturized device components. Compared to common interpolation methods, our qualitative and quantitative results indicate that a densely connected convolutional neural network can significantly yield higher PSNR and SSIM scores, resulting in super-resolved images of higher quality. \par In future work, we will focus on how super-resolved images, compared to low-resolution images, can be advantageous to clinical and research applications. For example, super-resolution images may be used as input to automated machine-learning based disease classification. 
%counting techniques, used to predict survival in cancer patients, may also see improvements when applied to CLE super-resolved images. 

\noindent \textbf{Acknowledgments}. Thanks to the NVIDIA Corporation for the donation of Titan X GPUs used in this research and to the Collaborative Health Research Projects (CHRP) for funding.

\bibliographystyle{splncs03}%IEEEbib}
\bibliography{refs}

% %
% % ---- Bibliography ----
% %
% \begin{thebibliography}{}
% %
% \bibitem[2017]{siegel2017colorectal}
% Siegel, Rebecca L and Miller, Kimberly D and Fedewa, Stacey A and Ahnen, Dennis J and Meester, Reinier GS and Barzi, Afsaneh and Jemal, Ahmedin:
% Colorectal cancer statistics. 

% \bibitem[1980]{2clar:eke}
% Clarke, F., Ekeland, I.:
% Nonlinear oscillations and
% boundary-value problems for Hamiltonian systems.
% Arch. Rat. Mech. Anal. 78, 315--333 (1982)

% \bibitem[1981]{2clar:eke:2}
% Clarke, F., Ekeland, I.:
% Solutions p\'{e}riodiques, du
% p\'{e}riode donn\'{e}e, des \'{e}quations hamiltoniennes.
% Note CRAS Paris 287, 1013--1015 (1978)

% \bibitem[1982]{2mich:tar}
% Michalek, R., Tarantello, G.:
% Subharmonic solutions with prescribed minimal
% period for nonautonomous Hamiltonian systems.
% J. Diff. Eq. 72, 28--55 (1988)

% \bibitem[1983]{2tar}
% Tarantello, G.:
% Subharmonic solutions for Hamiltonian
% systems via a $\bbbz_{p}$ pseudoindex theory.
% Annali di Matematica Pura (to appear)

% \bibitem[1985]{2rab}
% Rabinowitz, P.:
% On subharmonic solutions of a Hamiltonian system.
% Comm. Pure Appl. Math. 33, 609--633 (1980)

% \end{thebibliography}
\clearpage
% \addtocmark[2]{Author Index} % additional numbered TOC entry
% \renewcommand{\indexname}{Author Index}
% \printindex
% \clearpage
% \addtocmark[2]{Subject Index} % additional numbered TOC entry
% \markboth{Subject Index}{Subject Index}
% \renewcommand{\indexname}{Subject Index}
% \input{subjidx.tex}
\end{document}